# One-Leg Stance of Humanoid Robot using Active Balance Control

Tri Duc Tran[1], Anh Khoa Lanh Luu[1], Van Tu Duong[1,2], Huy Hung Nguyen[1,3] and Tan Tien Nguyen[1,*]
[1]*National Key Laboratory of Digital Control and System Engineering, Ho Chi Minh City University of Technology, VNU-HCM, Hochiminh city, Vietnam.*
[2]*The Department of Mechatronics, Ho Chi Minh City University of Technology, VNU-HCM, Hochiminh city, Vietnam.*
[3]*Faculty of Electronics and Telecommunication, Saigon University, Vietnam*
[*]Corresponding author. E-mail: nttien@hcmut.edu.vn

## Abstract

The task of self-balancing is one of the most important tasks when developing humanoid robots. This paper proposes a novel external balance mechanism for humanoid robot to maintain sideway balance. First, a dynamic model of the humanoid robot with balance mechanism and its simplified model are introduced. Secondly, a backstepping-based control method is utilized to split the system into two sub-systems. Then, a minimum observer-based controller is used to control the first sub-system. Since the second sub-system has unknown parameters, a model reference adaptive controller (MRAC) is used to control it. The proposed design divides the walking and balancing into two separated tasks, allowing the walking control can be executed independently of the balancing control. Furthermore, the use of the balance mechanism ensures the humanoid robot's hip movement does not exceed the threshold of a human when walking. Thus, making the overall pose of the humanoid robot looks more natural. An experiment is carried out on a commercial humanoid robot known as UXA-90 to evaluate the effectiveness of the proposed method.

## 1 Introduction

Ever since the old times, man has always been yearned for a replica of himself to serve in daily life. The idea of creating humanoid robots came to many great historical figures' minds. Leonardo da Vinci was believed to be the first man to have drawn a humanoid mechanism. At the beginning of 20$^{th}$ century, Elektro humanoid robot is made by the Westinghouse society (Chevallereau *et al.*, 2010). But it was not until the 1970s that an actual humanoid robot started to appear, Ichiro Kato from Waseda University has successfully built WABOT-1 – the first recognized humanoid robot in the world. Till now, many researchers have succeeded in developing their own humanoid robot with great capabilities (Kagami *et al.*, 2001; Sakagami *et al.*, 2002; Akachi *et al.*, 2005; Park *et al.*, 2007; Banerjee *et al.*, 2015; Tsagarakis *et al.*, 2017; Ko *et al.*, 2019). Despite that, the task of keeping balance on one leg is nowhere near good enough.

There have been many approaches to solve the problem, but it can be summarized into three strategies: ankle strategy, hip strategy and taking a step strategy (Stephens, 2007). The ankle strategy involves the adjustment of ankle torque to avoid falling. For example, Jun-Ho Oh et. al. (Kim, Park and Oh, 2007) utilized the inertial sensor feedback to adjust the ankle torque to make the humanoid robot staying upright. On the other hand, the hip strategy requires movements of different joints other than the ankle, which create the moment about the center of mass (CoM). Goswami et. al. (Goswami and Kallem, 2004) utilized a method similar to the hip strategy to keep the robot stable. The third strategy includes taking a step forward, backward, or sideway to maintain balance. Pratt et. al. (Pratt *et al.*, 2012) derived a velocity-based formulation and "Linear Inverted Pendulum Plus Flywheel Model" to formulate a "capture region", where the robot has to step to avoid falling. Hoffman (Hofmann, 2006) also studied balancing task in humanoid robot in his thesis, he argued that the control of horizontal motion of the CoM is effective in balancing the robot and analyzed the three strategies to accomplish this.

Another unique approach is to use an external mass to keep the balance. Jo et. al. (Jo and Mir-Nasiri, 2013) utilized three masses, two major static masses and one minor movable mass. However, the use of too many masses is redundant and makes the humanoid robot heavier. Also, the controller for the movable mass is the traditional PID, although this method is easy to implement, it does not guarantee robustness.

In our previous studies (Nguyen *et al.*, 2017; Van Tien *et al.*, 2017), UXA-90 is assumed to work in a perfect environment, which means the ground is perfectly flat with external forces and disturbances. While it can walk slowly with pre-computed trajectories, the humanoid robot lacks many features, one of them is to be able to stand on one leg.

In this paper, an external balance mechanism is built and crafted onto the humanoid robot's back. This mechanism allows the sideway balancing for the humanoid robot while standing on one leg. First, a simplified model for the humanoid robot with an external balance mechanism is derived. Secondly, a backstepping-based controller is implemented to decompose the aforementioned system into two sub-systems, then two controllers are proposed to control them. Since one of the roles of the external balance mechanism is to separate the walking task and the balancing task, a simple control method for the balance mechanism is crucial to reduce the computation load for the processor. For that reason, pole placement method is implemented, this method is not only simple but also guarantees the closed-loop stability. However, because pole placement method requires knowledge of all state variables, which is impossible to achieve, an observer to estimate the state variable is required. Therefore, the first controller is an observer-based method. The second controller is an MRAC based method due to the unknown parameters in the second sub-system. Finally, a simulation is carried out in MATLAB to evaluate the effectiveness of the proposed mechanism.

## 2 Humanoid robot and movable mass model

### 2.1 Full one-leg stance model

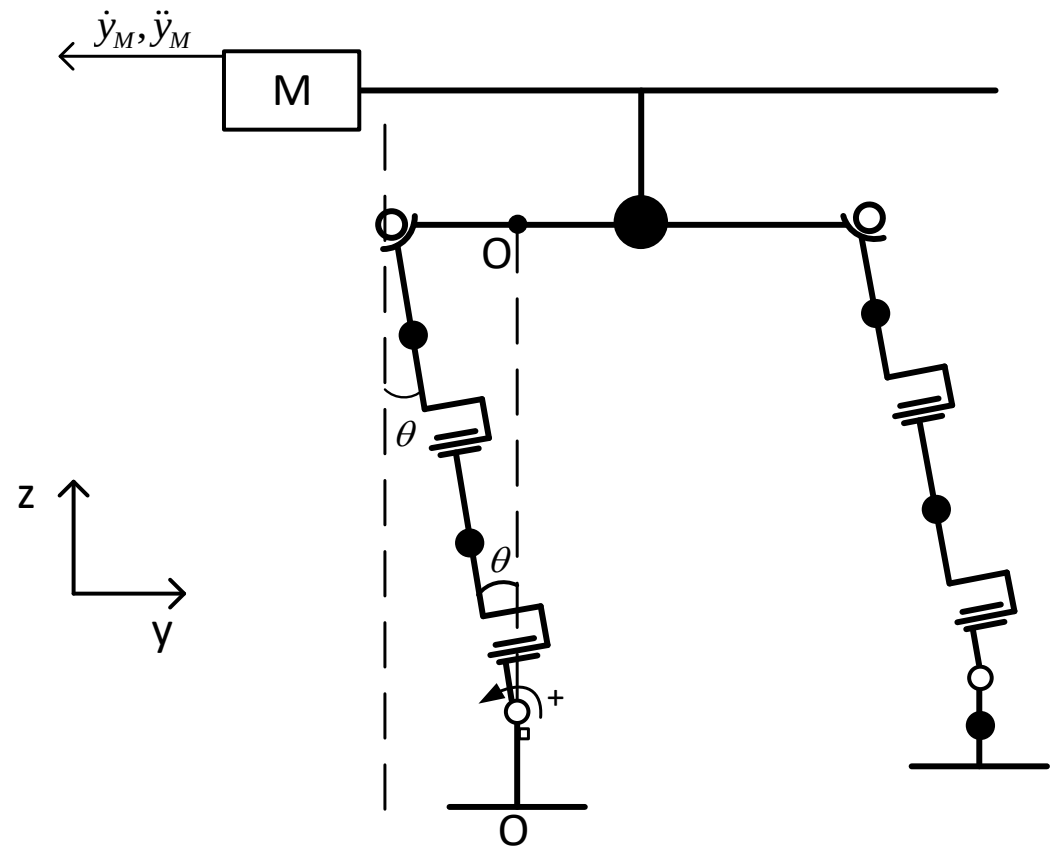

**Figure 1. Model of humanoid robot with balance mechanism**

The one-leg stance model for the robot with the balance mechanism attached to its back is shown in Fig. 1.

When moving or standing on one leg, man tends to rotate his hip and ankle slightly to shift his CoM toward the supporting feet, thus keeping the balance. According to some research (Chumanov, Wall-Scheffler and Heiderscheit, 2008; S., J. and B., 2019), when walking normally, the maximum value of hip adduction is 8.5° in women and 6.1° in men. To mimic this behavior naturally, the balance mechanism is mounted onto the back of UXA-90 to reduce its hip adduction when walking or standing on one leg.

We constrained the acceptable range for hip adduction is 6°~8.5°. Thus, the range of the moving mass to keep the system balance is calculated to be $6.24cm \rightarrow 10cm$.

### 2.2 Simplified one-leg stance model

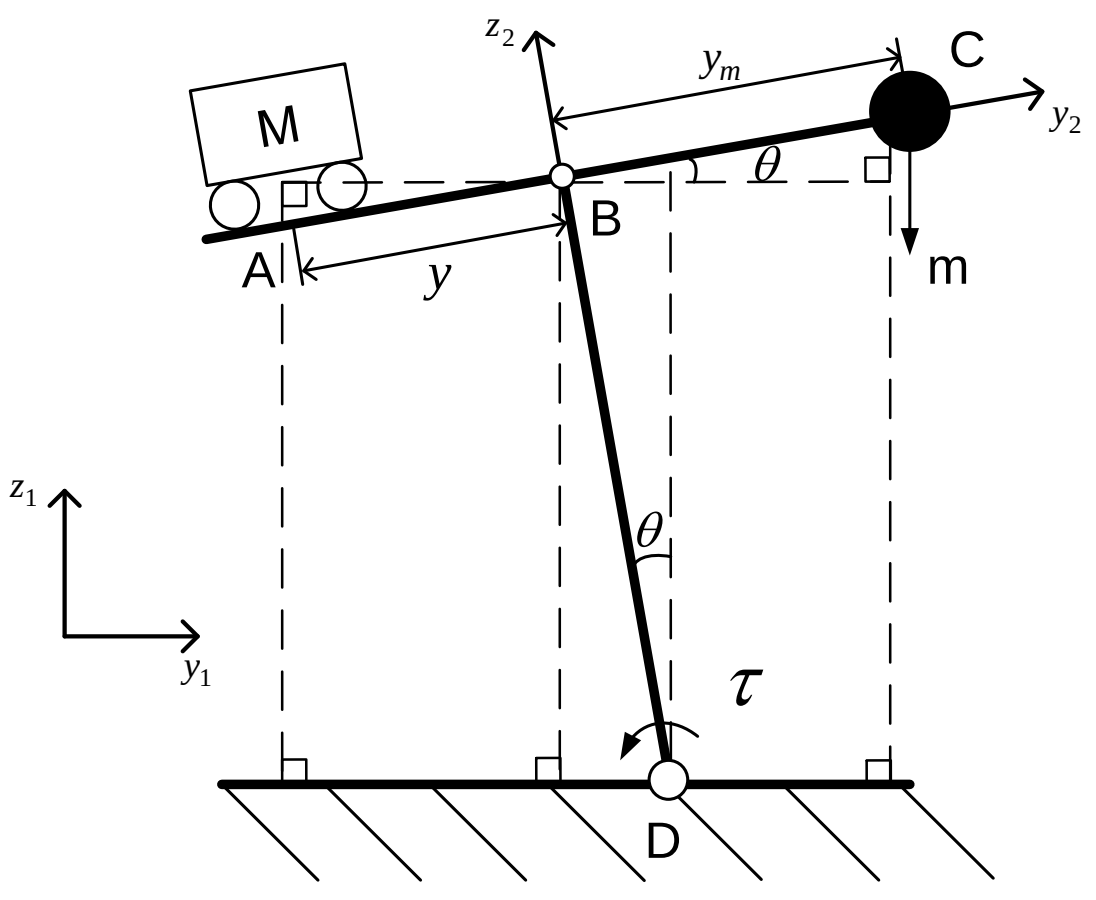

**Figure 2. Simplified one-leg stance model**

The simplified model for humanoid robot lifting one leg is described in Fig. 2.

In this model, the stance leg is BD and assumed to be massless. The mass of the body and swing leg are simplified as point mass $m$ at point C. The cart with mass $M$ moves horizontally along AC to balance the humanoid robot. The position of the cart and the simplified point mass are determined with respect to $z_2By_2$ axis, where B is the origin of this coordinate frame. The objective is to control the cart such that $\theta$ is as close to zero as possible.

First, let us derive a mathematical model for the cart-table model in Fig. 2.

By applying Euler-Lagrange, the dynamic equation for the one-leg stance is expressed as follows:

$$\tau = M\ddot{y}L - mg(y_m \cos\theta - L\sin\theta) + Mg(y\cos\theta + L\sin\theta) \tag{1}$$

where

$M$ is the external movable mass.

$y$ is the position of the external movable mass with respect to $z_2By_2$ axis.

$m$ is the total mass of the body and swing leg.

$y_m$ is the position of simplified mass point with respect to $z_2By_2$ axis.

$\theta$ is the tilting angle formed by the stance leg and the $z_1$ axis.

$L$ is the length of the stance leg.

The relationship between torque and moment of inertia is expressed as follows:

$$\tau = \ddot{\theta}(I_M + \sum I_m) \tag{2}$$

where

$\tau$ is the torque acting on point D.

$I_M$ is the movable mass's moment of inertia.

$I_m$ is the simplified mass's moment of inertia.

Substituting Eq. (2) into Eq. (1), it yields:

$$\ddot{\theta} = \frac{M\ddot{y}L - mg(y_m\cos\theta - L\sin\theta) + Mg(y\cos\theta + L\sin\theta)}{[M(L^2 + y^2) + m(L^2 + y_m^2)]} \tag{3}$$

The equilibrium point of Eq. (3) is reached when the following conditions are satisfied:

$$\theta = 0°, \dot{\theta} = 0, \ddot{\theta} = 0°, \dot{y} = 0, \ddot{y} = 0 \tag{4}$$

Recall in the previous section that the range of the movable mass $y$ is constrained to be $6.24cm \rightarrow 10cm$. Since the $L = 0.42m$ is the length of the humanoid robot's leg, therefore, we have $L^2 \gg y^2$, the $y^2$ term in Eq. (3) can be ignored. Thus, Eq. (3) can be rewritten as:

$$\ddot{\theta} = \frac{M\ddot{y}L - mg(y_m - L\theta) + Mg(y + L\theta)}{[ML^2 + m(L^2 + y_m^2)]} \tag{5}$$

By defining $\delta = [ML^2 + m(L^2 + y_m^2)]$, Eq. (5) becomes:

$$\ddot{\theta} = \frac{gL}{\delta}(M+m)\theta - \frac{mgy_m}{\delta} + \frac{Mg}{\delta}y + \frac{ML}{\delta}\ddot{y} \tag{6}$$

Substituting Eq. (4) into Eq. (6) yields:

$$\frac{Mg}{\delta}y - \frac{mg}{\delta}y_m = 0 \tag{7}$$

Eq. (7) shows the position relationship between the moving mass and the body's simplified mass when the humanoid robot is standing on one leg upright.

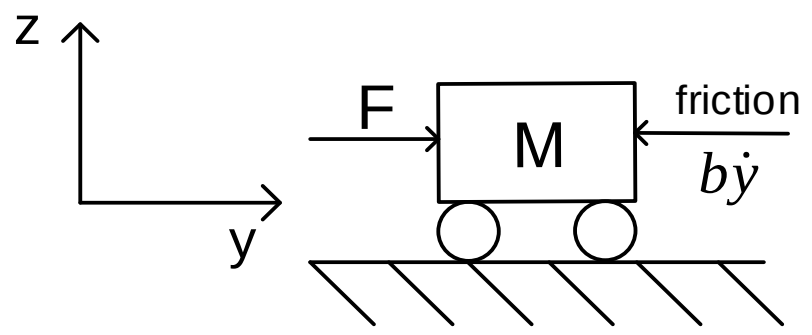

**Figure 3. The moving cart model**

To control the moving cart, a model of the moving cart shown in Fig. (3) is derived. According to Newton's second law, we have:

$$\ddot{y} = \frac{u_c}{M} - \frac{b}{M}\dot{y} \tag{8}$$

where

$u_c = F$ is the force applied to the cart.

$b$ is the friction coefficient.

Combining Eq. (7) and Eq. (8), it yields:

$$\begin{cases} \ddot{\theta} = \frac{gL}{\delta}(M+m)\theta - \frac{mgy_m}{\delta} + \frac{Mg}{\delta}y + \frac{ML}{\delta}\ddot{y} \\ \ddot{y} = \frac{u_c}{M} - \frac{b}{M}\dot{y} \end{cases} \tag{9}$$

Eq. (9) introduces the system modeling of the overall system with the combination of two subsystems, one-leg stance, and external balancing mechanism. In order to obtain the objective control, the proposed controller for the overall system is performed in two steps. First, the terms $y$ and $\ddot{y}$ are considered as the virtual control input of the first system in Eq. (9) to drive $\theta$ to zero. Then $y$ and $\ddot{y}$ are used as reference input for the second system of Eq. (9). Therefore, a backstepping technique is employed to build the proposed controller for the recursive structure system of Eq. (9).

## 3 Controller design

The block diagram of the overall control system is shown in Fig. 4.

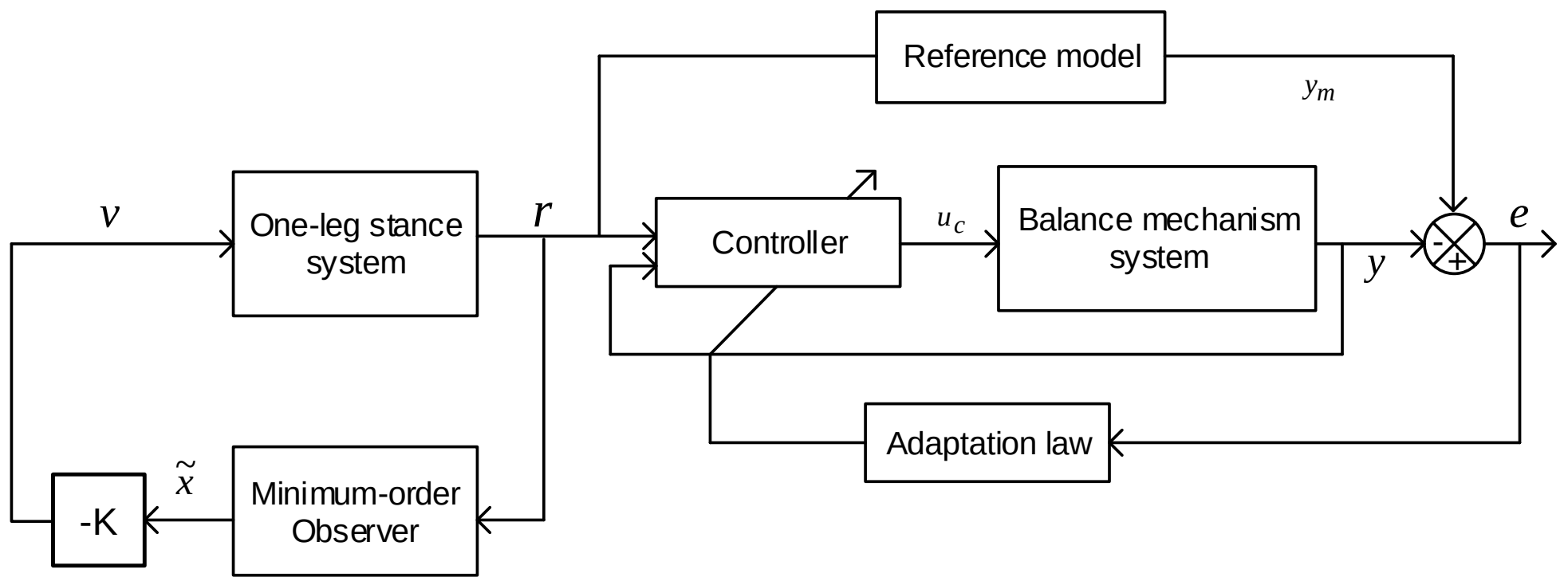

**Figure 4. Overall control system**

By defining the virtual control input $v = \frac{Mg}{\delta}y - \frac{mgy_m}{\delta}$, the first system of Eq. (9) becomes:

$$\ddot{\theta} = \frac{gL}{\delta}(M+m)\theta + v + \frac{L}{g}\ddot{v} \tag{10}$$

To eliminate the 2nd derivative of the virtual control input $v$. Let us define:

$$X_1 = \theta - \frac{L}{g}v, X_2 = \dot{\theta} - \frac{L}{g}\dot{v}, a = -\frac{gL}{\delta}(M+m), b = 1 + \frac{L^2}{\delta}(M+m) \tag{11}$$

Eq. (10) can be expressed in the state space:

$$\begin{bmatrix}\dot{X}_1\\ \dot{X}_2\end{bmatrix} = \begin{bmatrix}0 & 1\\ -a & 0\end{bmatrix}\begin{bmatrix}X_1\\ X_2\end{bmatrix} + \begin{bmatrix}0\\ b\end{bmatrix}v$$
$$Y = \begin{bmatrix}1 & 0\end{bmatrix}\begin{bmatrix}X_1\\ X_2\end{bmatrix} + \frac{L}{g}v \tag{12}$$

With $X_1 = \theta - \frac{L}{g}v$, recall Eq. (7) shows that $v \to 0$ as $t \to \infty$ when the system is at equilibrium state. Therefore, to drive the angle $\theta$ to zero, $X_1$ needs to converge to zero as $t$ tends to infinity.

Since only $X_1$ can be calculated through the measurement of $\theta$, $X_2$ needs to be estimated. Therefore, a Regulator System with Minimum-Order Observers is proposed to control the system.

**Step 1: Regulator with Minimum-order Observers design for the first system**

The first step in designing the Minimum-Order Observers is to divide the state vector of Eq. (12) into two parts, the measurable and unmeasurable state variable. The partitioned state of Eq. (12) is as follows:

$$\begin{bmatrix}\dot{x}_a\\ \dot{x}_b\end{bmatrix} = \begin{bmatrix}A_{aa} & A_{ab}\\ A_{ba} & A_{bb}\end{bmatrix}\begin{bmatrix}x_a\\ x_b\end{bmatrix} + \begin{bmatrix}B_a\\ B_b\end{bmatrix}v$$
$$Y = \begin{bmatrix}1 & 0\end{bmatrix}\begin{bmatrix}x_a\\ x_b\end{bmatrix} + \frac{L}{g}v \tag{13}$$

where $A_{aa} = 0, A_{ab} = 1, A_{ba} = -a, A_{bb} = 0, B_a = 0, B_b = 1.$

$x_a = X_1$ is measurable and $x_b = X_2$ is unmeasurable.

The plant and error for Minimum-Order Observers are given as (Ogata and Brewer, 1971):

$$\dot{\tilde{x}}_b = (A_{bb} - K_eA_{ab})\tilde{x}_b + A_{ba}x_a + B_bu + K_eA_{ab}x_b \tag{14}$$

$$\dot{e} = (A_{bb} - K_eA_{ab})x_b - \tilde{x}_b$$

where $x_b$ is the unmeasurable state variable.

$\tilde{x}_b$ is the estimated unmeasurable state variable.

$\dot{\tilde{x}}_b$ is the first derivative of the estimated unmeasurable state variable.

Substituting the value of $A_{aa}, A_{ab}, A_{ba}, A_{bb}, x_a, x_b$ into Eq. (14) it yields:

$$\dot{\tilde{X}}_2 = -K_e\tilde{X}_2 - aX_1 + u + K_eX_2$$
$$\dot{e} = -K_eX_2 - \tilde{X}_2 \tag{15}$$

where $K_e$ is the observer gain.

Assuming the control signal $v$ to be:

$$v = -\boldsymbol{K}\tilde{\boldsymbol{x}} \tag{16}$$

where $\boldsymbol{K}$ is the state feedback gain matrix.

The system with observed state feedback is shown in Fig. 5.

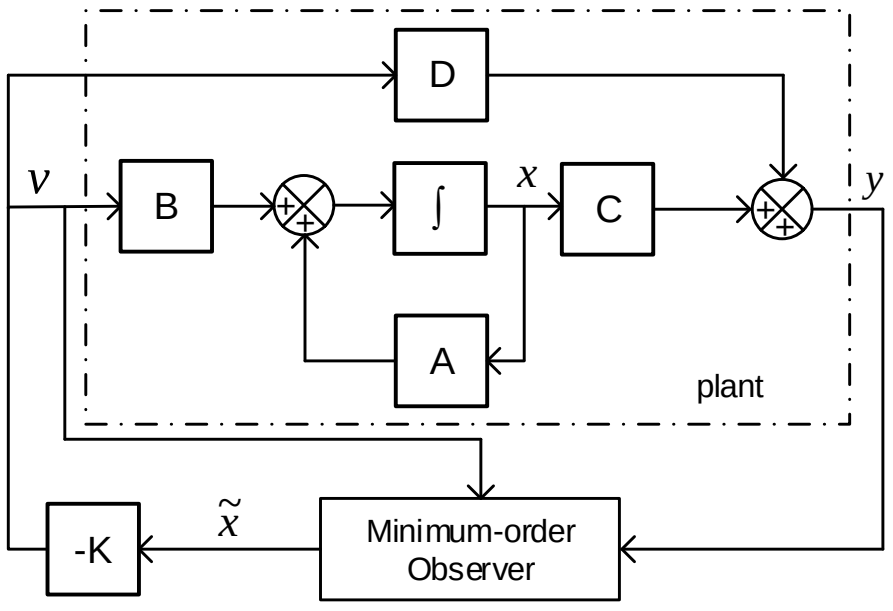

**Figure 5. Minimum-order Observer graph**

**Step 2: Adaptive control design for the second system**

Since only the friction coefficient $b$ is unknown, an MRAC is derived to control the second system in Eq. (9). By defining $\boldsymbol{X} = [y \quad \dot{y}]^T$, $\boldsymbol{A} = \begin{bmatrix} 0 & 1 \\ 0 & \frac{-b}{M} \end{bmatrix}$, $\boldsymbol{B} = \begin{bmatrix} 0 \\ \frac{1}{M} \end{bmatrix}$, $\Lambda = [1]$, the second system of Eq. (9) can be rewritten as follows:

$$\dot{\boldsymbol{X}} = \boldsymbol{A}\boldsymbol{X} + \boldsymbol{B}\Lambda u_c \tag{17}$$

Consider a reference model as below:

$$\dot{\boldsymbol{X}}_{\boldsymbol{m}} = \boldsymbol{A}_{\boldsymbol{m}}\boldsymbol{X}_{\boldsymbol{m}} + \boldsymbol{B}_{\boldsymbol{m}} r \tag{18}$$

where

$\boldsymbol{A}_{\boldsymbol{m}} = \begin{bmatrix} 0 & 1 \\ -1 & \frac{-b}{M} \end{bmatrix}$ is stable matrix.

$\boldsymbol{B}_{\boldsymbol{m}} = \begin{bmatrix} 0 \\ \frac{-b}{M} \end{bmatrix}$.

$r$ is bounded reference input vector.

The ideal control input and the tracking control input for the system in Eq. (17) is as follows:

$$u_{ideal} = \boldsymbol{K}_{\boldsymbol{x}}^{\boldsymbol{T}}\boldsymbol{X} + K_r r \tag{19}$$

$$u_c = \widehat{\boldsymbol{K}}_{\boldsymbol{x}}^{\boldsymbol{T}}\boldsymbol{X} + \widehat{K}_r r \tag{20}$$

where

$\boldsymbol{K}_{\boldsymbol{x}} \in \mathcal{R}^{2\times 1}$ is ideal feedback gain matrix.

$K_r$ is ideal feedforward gain parameter.

$\widehat{\boldsymbol{K}}_{\boldsymbol{x}} \in \mathcal{R}^{2\times 1}$ is estimated feedback gain matrix.

$\widehat{K}_r$ is estimated feedforward gain parameter.

Substituting Eq. (19) and Eq. (20) to Eq. (17), it yields:

$$\dot{\boldsymbol{X}} = (\boldsymbol{A} + \boldsymbol{B}\Lambda\boldsymbol{K}_{\boldsymbol{x}}^{\boldsymbol{T}})\boldsymbol{X} + \boldsymbol{B}\Lambda K_r r \tag{21}$$

$$\dot{\boldsymbol{X}} = \left(\boldsymbol{A} + \boldsymbol{B}\Lambda\widehat{\boldsymbol{K}}_{\boldsymbol{x}}^{\boldsymbol{T}}\right)\boldsymbol{X} + \boldsymbol{B}\Lambda\widehat{K}_r r \tag{22}$$

Comparing Eq. (21) and the reference model in Eq. (18), the ideal gain $\boldsymbol{K}_{\boldsymbol{x}}$ and $K_r$ must satisfy the matching condition:

$$\begin{cases} \boldsymbol{A} + \boldsymbol{B}\Lambda\boldsymbol{K}_{\boldsymbol{x}}^{\boldsymbol{T}} = \boldsymbol{A}_{\boldsymbol{m}} \\ \boldsymbol{B}\Lambda K_r = \boldsymbol{B}_{\boldsymbol{m}} \end{cases} \tag{23}$$

Using the matching condition in Eq. (23), Eq. (22) becomes:

$$\dot{\boldsymbol{X}} = \boldsymbol{A}_{\boldsymbol{m}}\boldsymbol{X} + \boldsymbol{B}_{\boldsymbol{m}} r + \boldsymbol{B}\Lambda\left(\widehat{\boldsymbol{K}}_{\boldsymbol{x}} - \boldsymbol{K}_{\boldsymbol{x}}\right)^T\boldsymbol{X} + \boldsymbol{B}\Lambda\left(\widehat{K}_r - K_r\right)r \tag{24}$$

The difference between the state of reference model in Eq. (18) and the state of the system in Eq. (17) is defined as:

$$\boldsymbol{e} = \boldsymbol{X} - \boldsymbol{X}_{\boldsymbol{m}} \tag{25}$$

Taking the 1st derivative of Eq. (25), it yields:

$$\dot{\boldsymbol{e}} = \dot{\boldsymbol{X}} - \dot{\boldsymbol{X}}_{\boldsymbol{m}} \tag{26}$$

Substituting Eq. (18) and Eq. (24) into Eq. (26), we have:

$$\dot{\boldsymbol{e}} = \boldsymbol{A}_{\boldsymbol{m}}(\boldsymbol{X} - \boldsymbol{X}_{\boldsymbol{m}}) + \boldsymbol{B}\Lambda\left(\widehat{\boldsymbol{K}}_{\boldsymbol{x}} - \boldsymbol{K}_{\boldsymbol{x}}\right)^T\boldsymbol{X} + \boldsymbol{B}\Lambda\left(\widehat{K}_r - K_r\right)r \tag{27}$$

Define the parameter estimation errors as follows:

$$\Delta \boldsymbol{K_x} = \boldsymbol{\hat{K}_x} - \boldsymbol{K_x}$$
$$\Delta K_r = \hat{K}_r - K_r \tag{28}$$

With Eq. (25) and Eq. (28), the error dynamic in Eq. (27) can be rewritten as:

$$\boldsymbol{\dot{e}} = \boldsymbol{A_m e} + \boldsymbol{B}\Lambda(\boldsymbol{\Delta K_x^T X} + \Delta K_r r) \tag{29}$$

Consider the Lyapunov function candidate for Eq. (29):

$$V(\boldsymbol{e}, \Delta \boldsymbol{K_x}, \Delta K_r) = \boldsymbol{e^T P e} + tr(\boldsymbol{\Delta K_x^T \Gamma_x^{-1}} \Delta \boldsymbol{K_x} \Lambda) + \Gamma_r^{-1} \Delta K_r^2 \Lambda \tag{30}$$

where

$\boldsymbol{\Gamma_x} = \boldsymbol{\Gamma_x^T} > 0$

$\Gamma_r > 0$

$\boldsymbol{P} = \boldsymbol{P^T} > \boldsymbol{0}$ and must satisfy the following algebraic Lyapunov equation:

$$\boldsymbol{P A_m} + \boldsymbol{A_m^T P} = -\boldsymbol{Q} \tag{31}$$

where $\boldsymbol{Q}$ is any symmetric positive definite matrix.

Taking the 1st derivative of Eq. (30) gives:

$$\dot{V}(\boldsymbol{e}, \Delta \boldsymbol{K_x}, \Delta K_r) = \boldsymbol{\dot{e}^T P e} + \boldsymbol{e^T P \dot{e}} + 2tr\left(\boldsymbol{\Delta K_x^T \Gamma_x^{-1} \dot{\hat{K}}_x} \Lambda\right) + 2\Gamma_r^{-1}\, \Delta K_r \dot{\hat{K}}_r \Lambda \tag{32}$$

Using Eq. (29), Eq. (32) can be rewritten as:

$$\dot{V}(\boldsymbol{e}, \Delta \boldsymbol{K_x}, \Delta K_r)$$
$$= \boldsymbol{e^T}(\boldsymbol{A_m^T P} + \boldsymbol{P A_m})\boldsymbol{e} + 2\boldsymbol{e^T P B}\Lambda(\boldsymbol{\Delta K_x^T X} + \Delta K_r r) + 2tr\left(\boldsymbol{\Delta K_x^T \Gamma_x^{-1} \dot{\hat{K}}_x}\Lambda\right) + 2\Gamma_r^{-1} \Delta K_r \dot{\hat{K}}_r \Lambda \tag{33}$$

Substituting Eq. (31) into Eq. (33), it yields:

$$\dot{V}(\boldsymbol{e}, \Delta \boldsymbol{K_x}, \Delta K_r)$$
$$= -\boldsymbol{e^T Q e} + \left[2\boldsymbol{e^T P B}\Lambda\boldsymbol{\Delta K_x^T X} + 2tr\left(\boldsymbol{\Delta K_x^T \Gamma_x^{-1} \dot{\hat{K}}_x}\Lambda\right)\right] + \left(2\boldsymbol{e^T P B}\Lambda \Delta K_r r + 2\Gamma_r^{-1}\, \Delta K_r \dot{\hat{K}}_r \Lambda\right) \tag{34}$$

Using the trace identity, we get:

$$\boldsymbol{e^T P B}\Lambda\boldsymbol{\Delta K_x^T X} = tr(\boldsymbol{\Delta K_x^T X e^T P B}\Lambda) \tag{35}$$

Substituting Eq. (35) into Eq. (34) yields:

$$\dot{V}(\boldsymbol{e}, \Delta \boldsymbol{K_x}, \Delta K_r)$$
$$= -\boldsymbol{e^T Q e} + 2tr\left[\boldsymbol{\Delta K_x^T}\left(\boldsymbol{X e^T P B} + \boldsymbol{\Gamma_x^{-1} \dot{\hat{K}}_x}\right)\Lambda\right] + 2\Lambda\Delta K_r\left(\boldsymbol{e^T P B} r + \Gamma_r^{-1}\dot{\hat{K}}_r\right) \tag{36}$$

The adaptive laws are chosen as follows:

$$\boldsymbol{\dot{\hat{K}}_x} = -\boldsymbol{\Gamma_x X e^T P B}$$
$$\dot{\hat{K}}_r = -\Gamma_r \boldsymbol{e^T P B} r \tag{37}$$

With Eq. (37), Eq. (36) becomes:

$$\dot{V}(\boldsymbol{e}, \Delta \boldsymbol{K_x}, \Delta K_r) = -\boldsymbol{e^T Q e} \le 0 \tag{38}$$

Hence, the tracking error $\boldsymbol{e}$ and the parameter estimation errors $\Delta \boldsymbol{K_x}$, $\Delta K_r$ are bounded.

Taking the 2nd derivative of Eq. (38), we have:

$$\ddot{V}(\boldsymbol{e}, \Delta \boldsymbol{K_x}, \Delta K_r) = -2\boldsymbol{e^T Q \dot{e}} \tag{39}$$

Eq. (39) showed that $\ddot{V}(t)$ is bounded, which means $\dot{V}(t)$ is uniformly continuous. According to Barbalat's Lemma, we have:

$$\lim_{t\to\infty} \dot{V}(t) = 0 \tag{40}$$

Thus the tracking error $\boldsymbol{e}$ will converge to zero as $t$ tends to infinity.

## 4 Simulations and results

The physical parameters for the model in Fig. 2 are given by Table I.

**Table I. Physical parameters for simplified one-leg stance model**

| Name | Length |
|---|---|
| $BD$ | $0.42m$ |
| $AB$ | $y$ |
| $BC$ | $y_m$ |

Since the response of the system depends on the initial condition, the chosen observer poles and controller poles, multiple simulations are carried out in MATLAB to analyze the system's response. First, by regulating $\theta$ to zero in the first system of Eq. (9), the trajectory of the moving mass is obtained. The aforementioned trajectory is then used as reference for the second system. The sampling time is 0.01 second.

- **Scenario 1: The hip adduction of humanoid robot is $6°$, the position of the simplified point mass is calculated to be $y_m \approx 0.049$.**

  The two closed-loop poles and observer pole are $\mu_1 = -4.5, \mu_2 = -4.5, s = -10$ respectively.

  The adaptive gains are chosen as follows:

$$\mathbf{\Gamma}_x = \begin{bmatrix} 10000 & 0 \\ 0 & 2000 \end{bmatrix}$$

$$\Gamma_r = 10$$

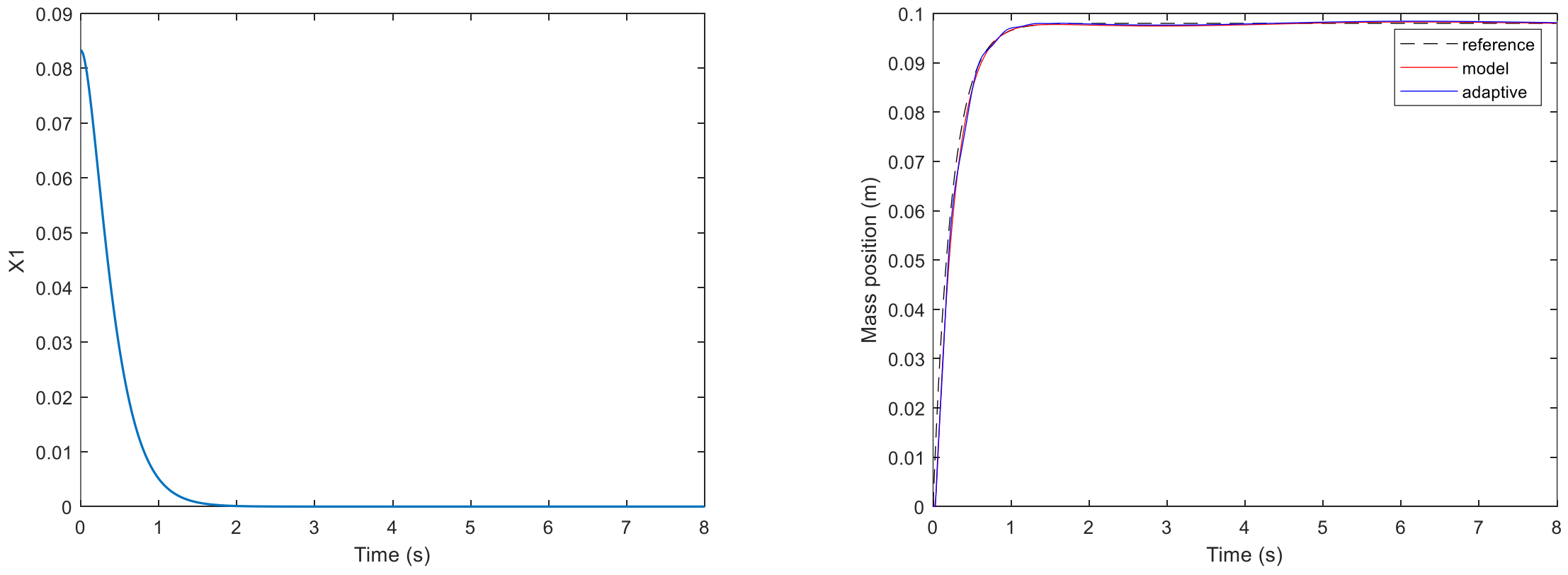

**Figure 6. $X_1$ response and the tracking of the position of movable mass when $y_m = 0.049m$**

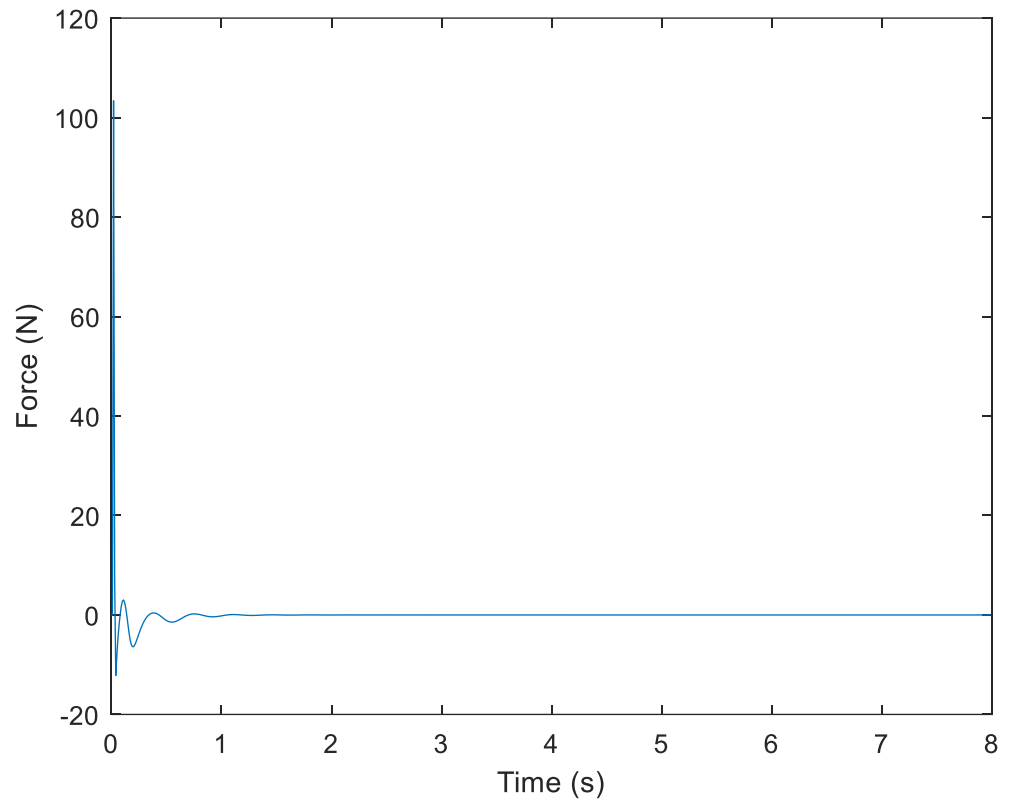

**Figure 7. Control signal $u_c$ for the movable mass**

- **Scenario 2: The hip adduction of humanoid robot is $7.3°$, the position of the simplified point mass is calculated to be $y_m \approx 0.04$.**

  The two closed-loop poles and observer pole are $\mu_1 = -4.5, \mu_2 = -4.5, s = -10$ respectively.

  The adaptive gains are chosen as follows:

$$\mathbf{\Gamma}_x = \begin{bmatrix} 5000 & 0 \\ 0 & 4000 \end{bmatrix}$$

$$\Gamma_r = 14$$

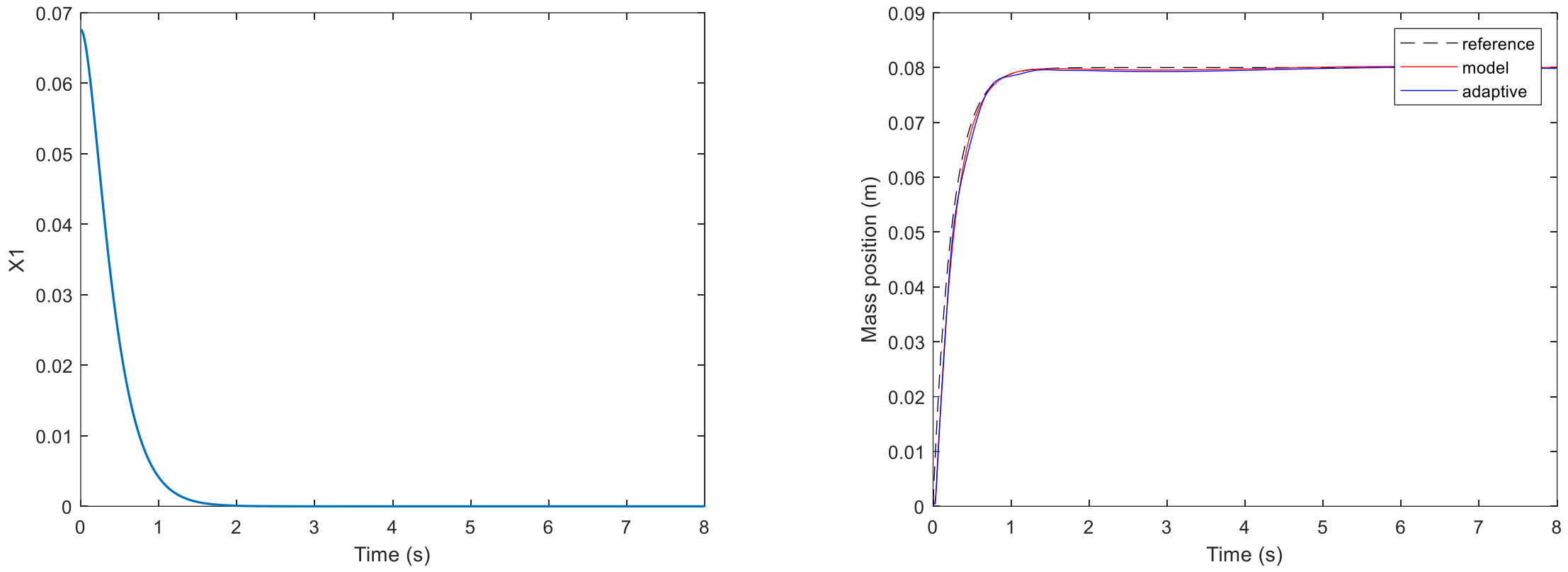

**Figure 8. $X_1$ response and the tracking of the position of movable mass when $y_m = 0.04m$**

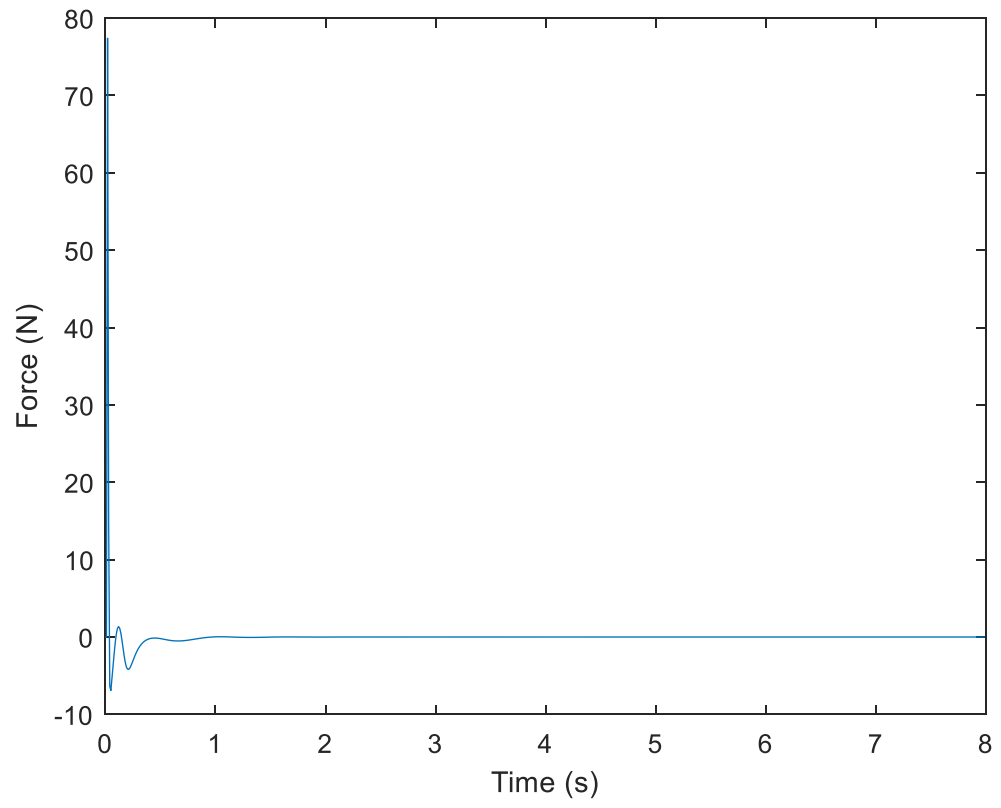

**Figure 9. Control signal $u_c$ for the movable mass**

- **Scenario 3: The hip adduction of humanoid robot is 8°, the position of the simplified point mass is calculated to be $y_m \approx 0.035m$**

  The two closed-loop poles and observer pole are $\mu_1 = -7, \mu_2 = -7, s = -14$ respectively.
  The adaptive gains are chosen as follows:

$$\boldsymbol{\Gamma}_x = \begin{bmatrix} 7000 & 0 \\ 0 & 700 \end{bmatrix}$$
$$\Gamma_r = 6.3$$

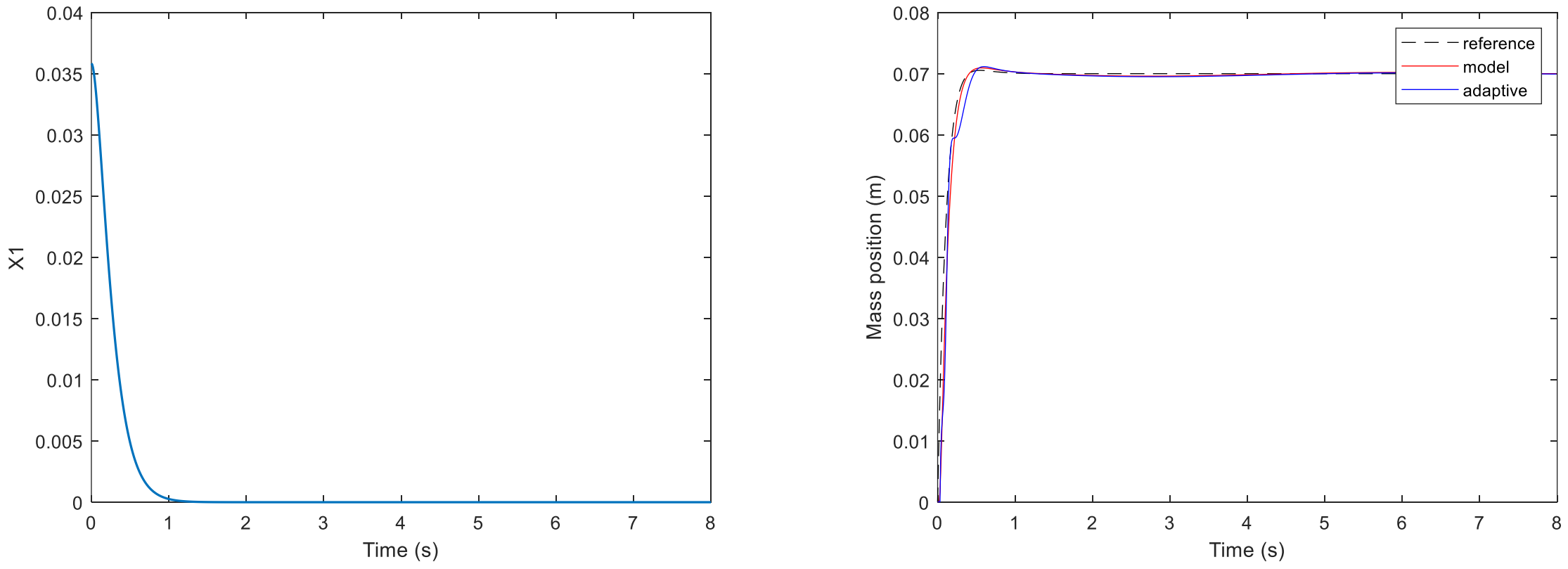

**Figure 10. $X_1$ response and the tracking of the position of movable mass when $y_m = 0.035m$**

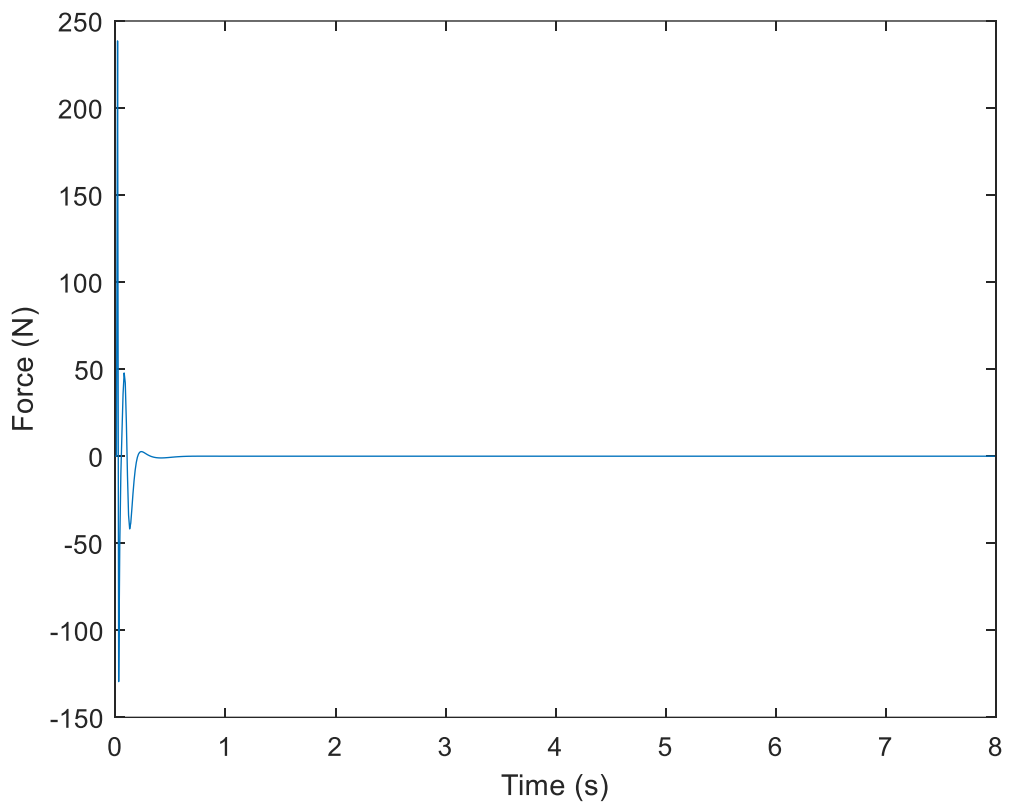

**Figure 11. Control signal $u_c$ for the movable mass**

- **Scenario 4: The hip adduction of humanoid robot is $8.5°$, the position of the simplified point mass is calculated to be $y_m \approx 0.03m$**

  The two closed-loop poles and observer pole are $\mu_1 = -2, \mu_2 = -2, s = -4$ respectively.
  
  The adaptive gains are chosen as follows:

$$\boldsymbol{\Gamma}_x = \begin{bmatrix} 7000 & 0 \\ 0 & 5000 \end{bmatrix}$$
$$\Gamma_r = 100$$

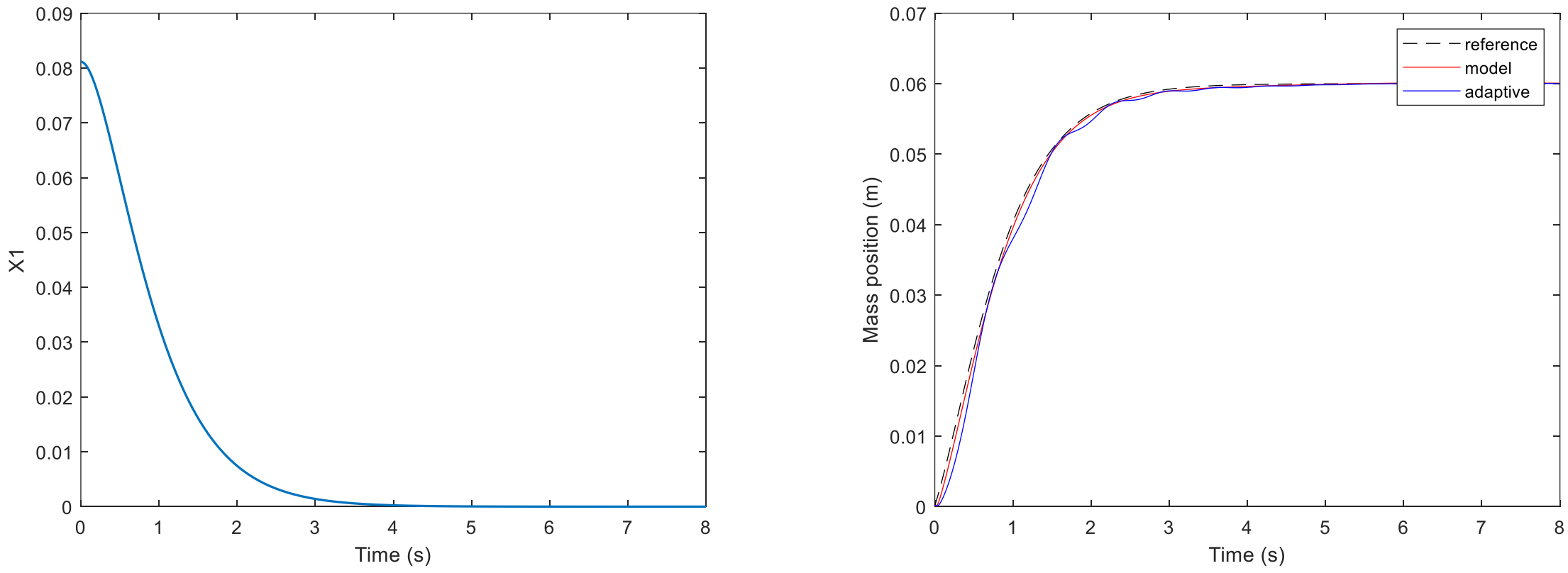

**Figure 12. $X_1$ response and the tracking of the position of movable mass when $y_m = 0.03m$**

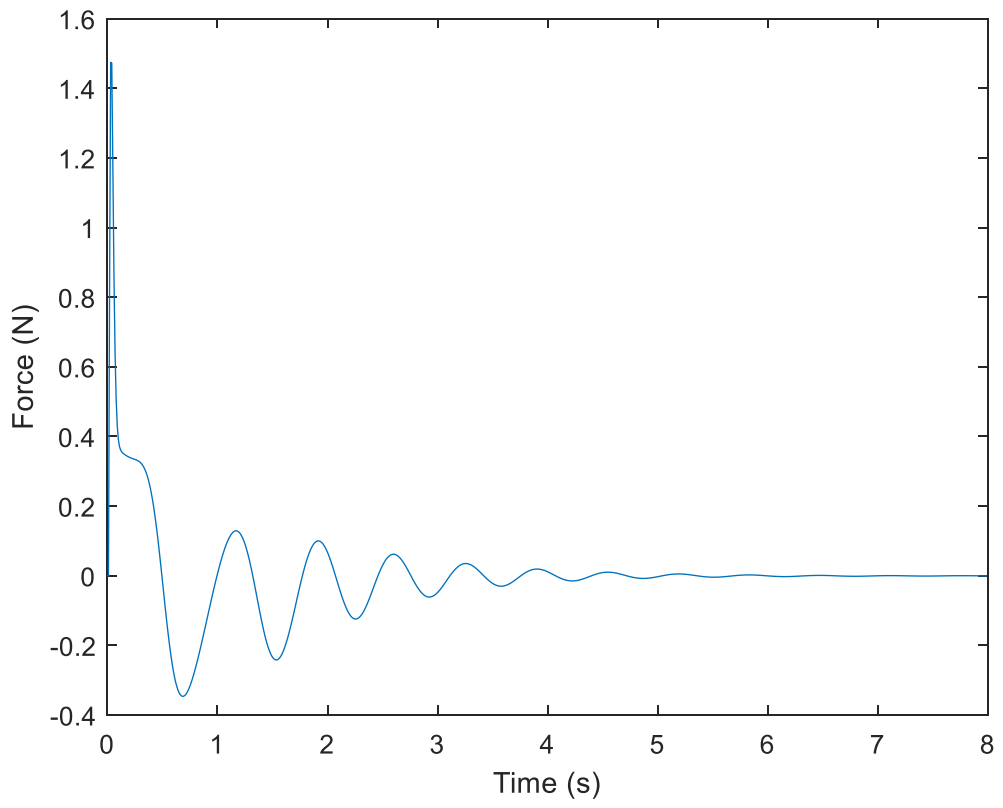

**Figure 13. Control signal $u_c$ for the movable mass**

In general, the controller drove the state variable $\boldsymbol{X_1}(t)$ in the one-leg stance system (first system) to zero in every simulation as shown in the first image of Fig 6, 8, 10, 12 which means $\theta$ was regulated to zero. The movable mass system (second system) also tracked the desired position closely as shown in the second image. Thus the controller satisfies the goal.

The simulations show that the choice of the adaptive gains $\boldsymbol{\Gamma}_x$, $\Gamma_r$ varies with different positions of $y_m$ and the poles. The further the poles are placed, the faster the $\theta$ in the first system is regulated to zero. However, because the response time of the first system decreases, the second system suffers from a slight overshoot as shown in the second image in Fig. 10. Since the adaptive gain $\Gamma_r$ is affected directly by the reference input of the second system, if the first system responds too fast, the second system will be extremely sensitive to a slight change in $\Gamma_r$, making it difficult to tune the controller.

## 5 Conclusions

This paper proposed a novel external balancing mechanism for humanoid robot to maintain sideway balance when standing on one leg. The backstepping method is employed to split the system into two subsystems, then a minimum-order observer controller and MRAC are utilized to control the two subsystems. The MRAC guarantees the system stability even when subjected to unknown parameters. The simulation results show that the MRAC successfully tracked the desired mass trajectory and the $\theta$ is driven to zero, satisfying the control objective. In the future, experiment will be made on humanoid robot UXA-90 to evaluate the effectiveness of the proposed method. Furthermore, UXA-90 is still subjected to disturbances when walking or standing still. Therefore, a new controller utilizing the external balance mechanism to reject disturbances will be researched in depth.

## Acknowledgments

This research is funded by Vietnam National University HoChiMinh City (VNU-HCM) under grant number B2019-20-09 and TX2021-20B-01.